\documentclass[conference]{IEEEtran}
\ifCLASSINFOpdf
\else
\fi
\usepackage{multicol}
\usepackage{graphicx}
\usepackage{amsmath}  
\usepackage[export]{adjustbox}

\begin{document}
%
\title{A Human Mixed Strategy Approach to Deep Reinforcement Learning}

\author{\IEEEauthorblockN{Ngoc Duy Nguyen, Saeid Nahavandi, and Thanh Nguyen}
\IEEEauthorblockA{Institute for Intelligent Systems Research and Innovation\\Deakin University, Waurn Ponds Campus, Geelong, VIC, 3216, Australia\\E-mails: \{duy.nguyen, saeid.nahavandi, thanh.nguyen\}@deakin.edu.au}
}


%


\maketitle

\begin{abstract}
In 2015, Google's Deepmind announced an advancement in creating an autonomous agent based on deep reinforcement learning (DRL) that could beat a professional player in a series of 49 Atari games. However, the current manifestation of DRL is still immature, and has significant drawbacks. One of DRL's imperfections is its lack of ``exploration" during the training process, especially when working with high-dimensional problems. In this paper, we propose a mixed strategy approach that mimics behaviors of human when interacting with environment, and create a ``thinking" agent that allows for more efficient exploration in the DRL training process. The simulation results based on the Breakout game show that our scheme achieves a higher probability of obtaining a maximum score than does the baseline DRL algorithm, i.e., the asynchronous advantage actor-critic method. The proposed scheme therefore can be applied effectively to solving a complicated task in a real-world application.
\end{abstract}


%
\IEEEpeerreviewmaketitle

\section{Introduction}
\label{sec:1}
Recent advances in \emph{deep learning} \cite{0} have made \emph{reinforcement learning} (RL) \cite{1} a possible solution for creating an agent that can mimic human behaviors \cite{2,2b,3,3b}. In 2015, for the first time, Mnih \emph{et al.} \cite{4} succeeded in training an agent to surpass human performance on playing Atari games. By employing a \emph{convolutional} layer \cite{4a}, the agent directly perceives the environment's state in the form of a graphical representation. Furthermore, the agent responds with a proper action for each perceived state to maximize the long-term reward. Specifically, Mnih \emph{et al.} \cite{4} created a novel structure, named \emph{deep Q-network} (DQN), which simulated the human brain to take decisive actions in a series of 49 Atari games. As a result, DQN initiates a new research branch of machine learning called deep RL that has recently attracted considerable research attention.

Since 2015, there have been extensive improvements to DQN. However, most of these variants substantially modify DQN structure in some aspects to fill the gap. For example, Hasselt \cite{5,6} explored the idea of \emph{double Q-learning} to stabilize the convergence of DQN; Schaul \emph{et al.} \cite{7} reduced correlated samples by assigning a priority to each transition in experience replay using \emph{temporal-difference} (TD) error; Wang \emph{et al.} \cite{8} adjusted DQN's policy network to forward the agent's attention to only the important regions of the game; and Hausknecht and Stone \cite{9} added a recurrent layer to DQN to prolong the agent's memory. In 2016, Mnih \emph{et al.} \cite{10} proposed another asynchronous method of deep RL called \emph{asynchronous advantage actor-critic} (A3C). A3C combines \emph{actor-critic} architecture \cite{11}, advantage function, and multithreading to drastically improve DQN in both perspectives: training speed and score achievement. Therefore, in this paper, we compare our proposed scheme with A3C, which is considered as the baseline deep RL algorithm.

\begin{figure}[!t]
\centering
\includegraphics[width=3.4in]{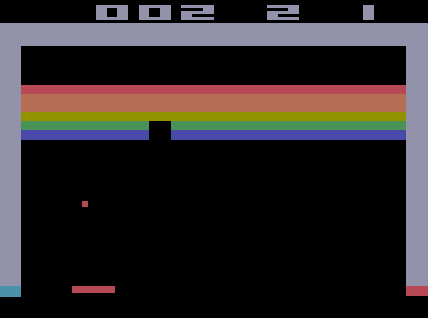}
\caption{A Breakout's gameplay using Arcade Learning Environment \cite{12}.}
\label{fig:1} 
\end{figure}

Our proposed scheme is initially motivated by the human brain's activities while playing the game. The human brain naturally divides a complicated task into a series of smaller and easier functional missions. This strategy -- divide and conquer -- is shown in everyday human activities. In this paper, we integrate this strategy into deep RL to create a human-like agent. Furthermore, we demonstrate our proposed scheme in the Breakout game using the \emph{Arcade Learning Environment} \cite{12}. In Breakout, the player controls the red paddle to the left or the right so that the paddle catches the ball falling, as shown in Fig.~\ref{fig:1}. When the ball touches the paddle, it bounces back and breaks the bricks at the top of the screen. The goal is to break as many bricks as possible and to keep the ball above the paddle at all times. If the paddle misses the ball, the player loses a ball's life. If the player cannot catch the ball five consecutive times, the game is over.

\begin{figure*}
\begin{multicols}{2}
    \includegraphics[width=3.4in]{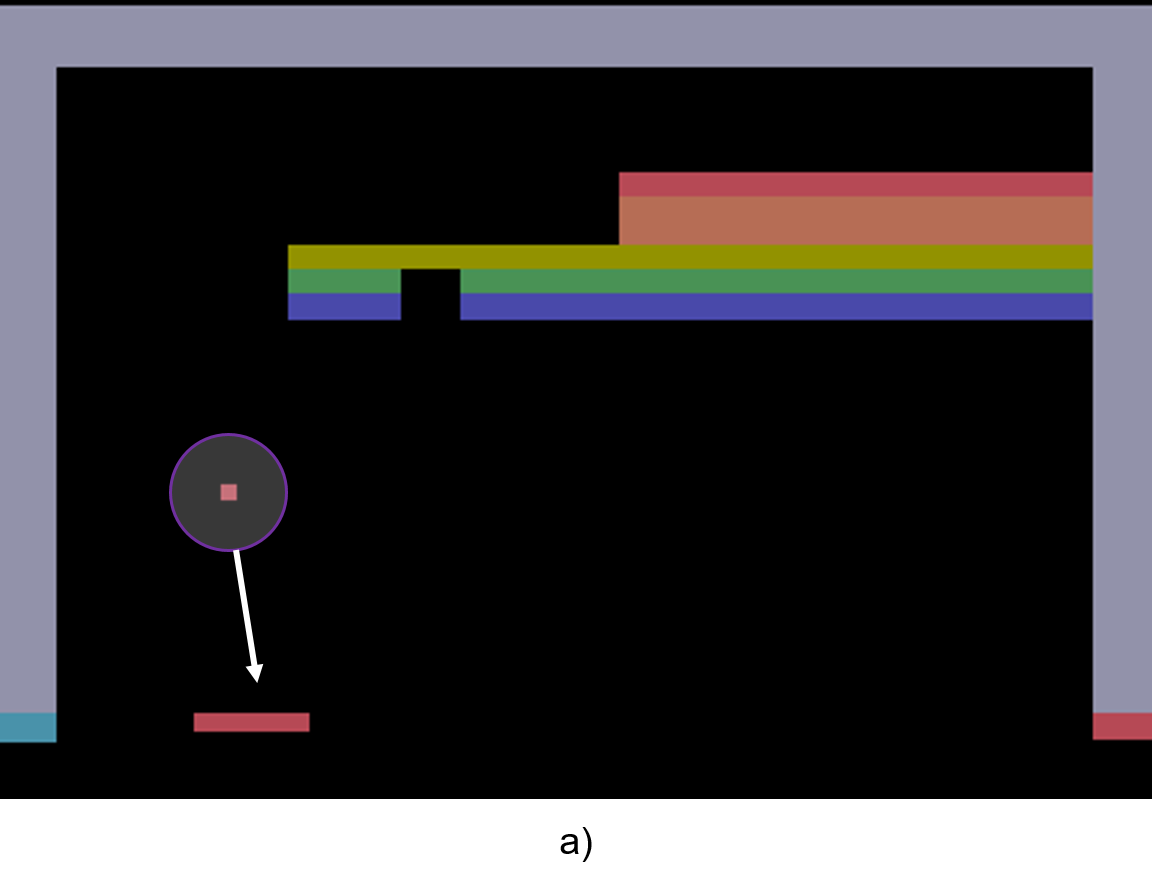}\par 
    \includegraphics[width=3.4in]{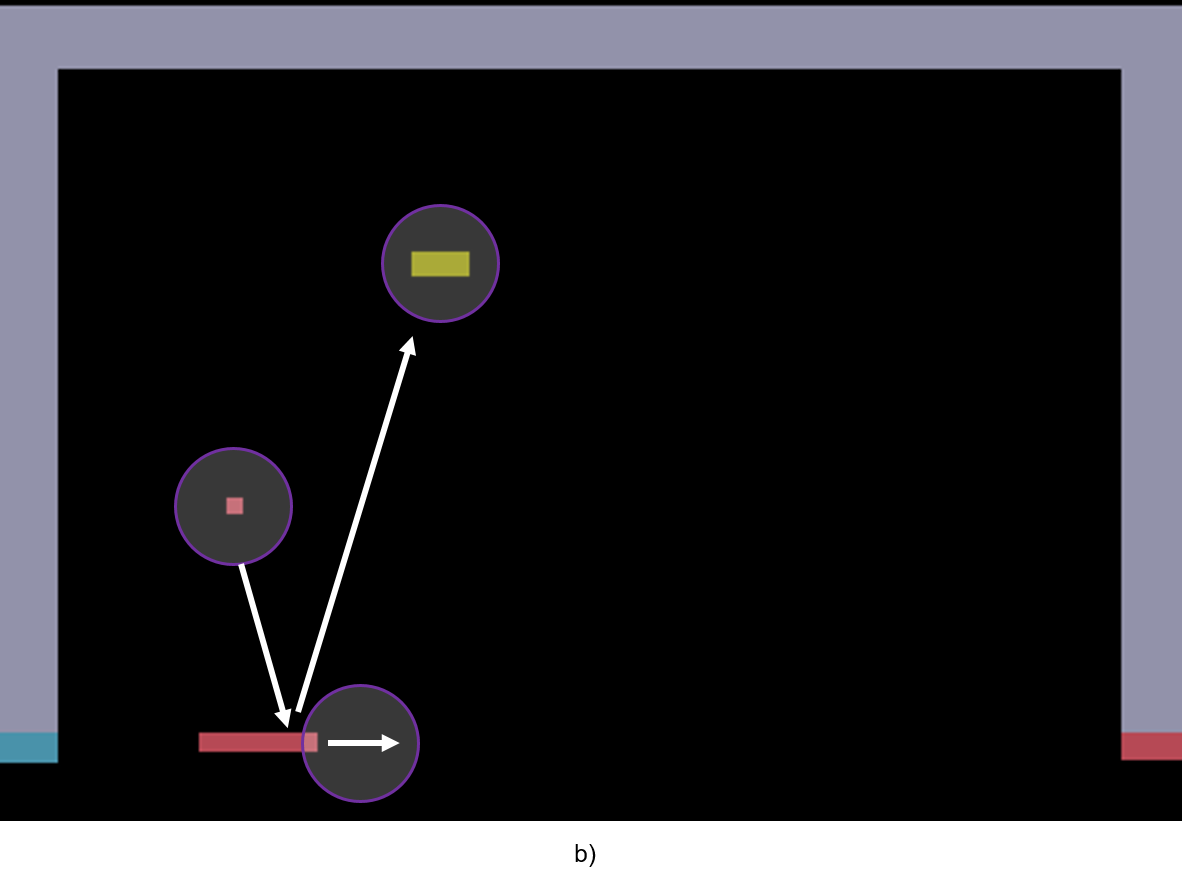}\par 
    \end{multicols}
\caption{Human strategy when playing Breakout. a) If there are full bricks in the game, we only focus on the ball. b) If there are few bricks left, we focus on the ball, the speed and the direction of the paddle, and the position of the brick.}
\label{fig:2} 
\end{figure*}

Fig.~\ref{fig:2} illustrates in detail a human strategy to achieve a high score when playing Breakout. In the beginning of gameplay, all bricks remain at the top of the screen, and the strategy is to focus only on the ball motion to assure a successful catch, as the probability of breaking a brick in this case is high. Gradually, the player focuses attention on the bricks towards the end of the game, when there are few bricks left. This scenario is more complicated, because we not only focus on the ball, but also its speed and the position of the brick. In Atari games, deep RL algorithms often perceive an environment's state as a whole. Therefore, any unimportant changes in the environment may cause unintended noise, and hence may degrade the algorithm's performance. In this paper, we suggest an approach that partially eliminates this drawback, reduces input data density, and encourages further exploration. 

In summary, the paper brings the following key contributions:

\begin{itemize}
\item We took a first step to integrate a human strategy into deep RL. Moreover, the proposed scheme is \emph{general} in the sense that it can be used with any deep RL algorithm. In this paper, we examine our divide-and-conquer strategy for A3C, a state-of-the-art deep RL algorithm.

\item Although we only demonstrate our approach using the Breakout game because of the limited scope of this paper, our proposed scheme is \emph{extendable}, \emph{i.e.}, it can be employed in any games as well as in real-world applications.

\item We provide helpful guidelines to solve a complicated task by using a divide-and-conquer strategy. Specifically, we \emph{divide} a complicated task $T$ into smaller tasks $t_i$, then use different strategies to \emph{conquer} each task $t_i$. Finally, we combine multiple policies using a generalized version of the stochastic $\epsilon$-greedy rule to produce a single mixed strategy policy. Therefore, the resulting agent becomes more human-like and flexible with the stochastic environment.

\end{itemize}

The paper is organized as follows: the next section summarizes variants of the deep RL algorithm; Section \ref{sec:3} illustrates our proposed scheme; Section \ref{sec:4} shows our simulation results; and Section \ref{sec:5} concludes our work.

\section{Related Work}
\label{sec:2}

As mentioned in Section \ref{sec:1}, DQN \cite{4} is the first successful attempt to combine deep learning with RL. The key of DQN is the utilization of a neural network to approximate optimal value function by minimizing the following loss function:

\begin{equation}
L_{\theta} \sim E\left[(r + \gamma \max_{a'}Q(s',a';\theta') - Q(s,a;\theta))^2\right],
\label{eq:1}
\end{equation}

\noindent
where $\theta$ and $\theta'$ represent the parameters of the estimation network and target network, respectively. To break the correlation between samples and to stabilize the convergence, Mnih \emph{et al.} \cite{4} introduced an \emph{experience replay} that is used to store history samples and a \emph{target network} that is updated asynchronously for every $N$ steps from the estimation network. Although DQN can solve a challenging problem in RL literature, it still has drawbacks, and has been improving since its inception in 2015. At first, Hasselt \emph{et al.} \cite{6} proposed a \emph{double deep Q-network} (DDQN) to reduce the overfitting problem in Q-learning by separating action evaluation from selection. In other words, the loss function (\ref{eq:1}) is replaced by the following function:

\begin{equation*}
L_{\theta} \sim E\left[(r + \gamma Q(s', \arg \max_{a'}Q(s',a';\theta); \theta')  - Q(s,a;\theta))^2\right].
\label{eq:2}
\end{equation*}

To promote ``rare" samples, Schaul \emph{et al.} \cite{7} proposed a \emph{prioritized experience replay} that assigns each sample in the experience replay a priority number based on its TD-error. Finally, Wang \emph{et al.} \cite{8} introduced a \emph{dueling network} architecture that breaks down a Q-value $Q(s,a)$ into state value $V(s)$ and advantage action value $A(s,a)$, as below:
 
\begin{equation*}
Q(s,a) \sim V(s) + \left(A(s,a) - \frac{1}{|\mathcal{A}|}\sum_{a'}A(s,a')\right),
\label{eq:3}
\end{equation*}

\noindent
where $|\mathcal{A}|$ denotes the number of possible actions. The dueling network helps to stabilize the policy network, especially in environments with sparse rewards.

Another major drawback of DQN is training time. DQN requires a training time of 7--8 days to surpass human performance in each Atari game. Therefore, in 2016, Mnih \emph{et al.} \cite{10} introduced an asynchronous version of DQN as well as of A3C. The simulation shows that A3C drastically speeds up the training process to only 1--3 days on CPU compared to DQN. Therefore, A3C becomes a baseline approach for deep RL. In this paper, we use A3C as a benchmark algorithm for comparisons with our proposed scheme.

\section{Proposed Scheme}
\label{sec:3}

\begin{figure}[!t]
\centering
\includegraphics[width=3.4in]{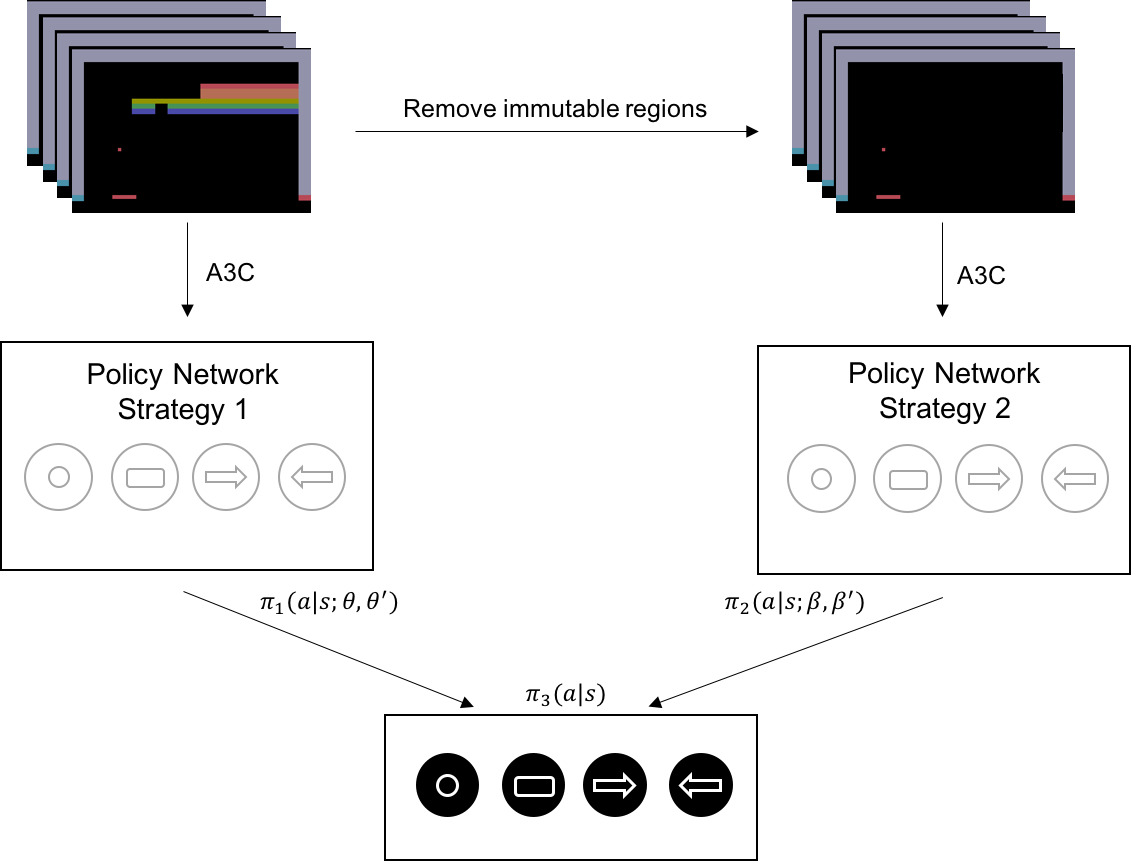}
\caption{A divide and conquer approach with two strategies using A3C.}
\label{fig:3} 
\end{figure}

As mentioned above, we use A3C as the base deep RL algorithm to integrate our divide-and-conquer strategy. Note that A3C uses \emph{actor-critic architecture}, which was proposed by Konda and Tsitsiklis \cite{13}. Therefore, there exist two policies in A3C, one for the actor network and another for the critic network. The actor network, parameterized by $\theta$, represents a stochastic policy, $\pi(a_i|s;\theta)$. It perceives state $s$ as input and produces probabilities for all possible actions $a_i$ as output. On the other hand, the critic network, parameterized by $\theta'$, represents a value function at state $s$, $V(s;\theta')$. The overall objective of A3C is to minimize the following loss function \cite{10}:

\begin{equation}
\begin{split}
L_{\theta, \theta'} \sim \log(\pi(a_t|s_t;\theta)) (R_t - V(s_t;\theta')) + \beta H(\pi(s_t;\theta)),
\end{split}
\label{eq:4}
\end{equation}

\noindent
where $H$ denotes the entropy function and:

\begin{equation*}
\begin{cases}
R_t = \sum_{i=0}^{k}\gamma^i r_{t+i} + \gamma^{k+1} V(s_{t+k+1};\theta')(1-T(s_{t+k+1}))\\
T(s_t) = 0 \text{ if } s_t \text{ is not a terminal state} \\
T(s_t) = 1 \text{ if } s_t \text{ is a terminal state} \\
\end{cases}
\label{eq:5}
\end{equation*}

\begin{figure}[!t]
\centering
\includegraphics[width=3.4in]{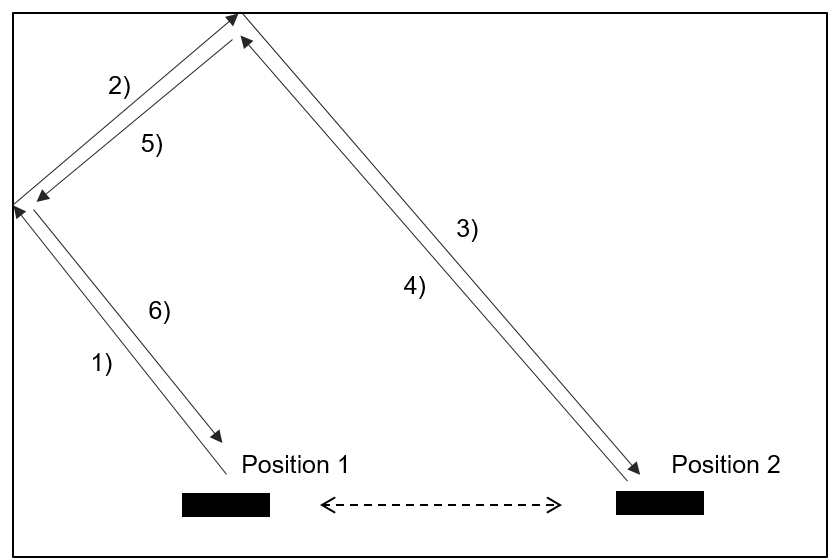}
\caption{A ``local stuck" phenomenon in Breakout due to lack of exploration.}
\label{fig:4} 
\end{figure}

\noindent
The entropy regularization term $\beta H(\pi(s_t;\theta)$ in (\ref{eq:4}) is used to encourage exploration in the training process. In practice, the actor network and critic network often share parameters in convolutional layers. Therefore, we can assume that the actor and critic network are actually a unique network with two output layers, denoted as $\pi(\theta, \theta')$. 

Based on A3C, our proposed scheme (Fig.~\ref{fig:3}) can be described in the following three steps:

1) First, we train a policy network $\pi_1(\theta, \theta')$ to learn Breakout using A3C. The state $s$ of the environment (a history of four frames) is converted to grayscale and fed directly to $\pi_1$. In this way, the policy $\pi_1$ is trained to learn all aspects of the game, including the position of the paddle, the ball motion, and the regions of the bricks. As explained earlier, any unimportant changes in $s$ may degrade the performance of the algorithm. Therefore, the policy $\pi_1$ represents a human strategy, as shown in Fig.~\ref{fig:2}b.

2) Second, we use A3C to train a second network $\pi_2(\beta, \beta')$. We remove all immutable objects in the input state $s$ before feeding it to $\pi_2$, and give a negative reward for any life lost. In our implementation, we blacken all immutable objects in state $s$. This policy is a life safeguard. It focuses only on the ball, and continues to catch the ball regardless of the presence or absence of bricks. Apparently, this strategy is only suitable in the beginning of the game, as shown in Fig.~\ref{fig:2}a. The use of this pure strategy can lead to a negative effect, which we name ``local stuck". This phenomenon occurs when the gameplay is stuck in an infinite loop. In Fig.~\ref{fig:4}, for example, the paddle moves only between two different positions, which leads to a loop circle of ball motion $1)\rightarrow2)\rightarrow3)\rightarrow4)\rightarrow5)\rightarrow6)\rightarrow1) ... $. Therefore, in this way, the game becomes stuck. This phenomenon occurs due to the lack of exploration in the training process. It also occurs in policy $\pi_1$, but with less frequency. The ``local stuck" phenomenon can be observed easily here\footnotemark.

\footnotetext{https://youtu.be/gbcdPSQP4XI.}

3) Finally, we combine $\pi_1$ and $\pi_2$ to create a stochastic policy $\pi_3$ using the following generalized version of the $\epsilon$-greedy rule with two strategies:

\begin{equation*}
    \pi_3(a|s) = \frac{\epsilon}{|\mathcal{A}|} + \alpha(1-\epsilon)\pi_1(a|s) + (1-\alpha)(1-\epsilon)\pi_2(a|s), 
\label{eq:6}
\end{equation*}

\noindent
where $0 \leq \alpha,\epsilon \leq 1$ and $\forall a \in \mathcal{A}$. Because $\sum_{i}\pi_1(a_i|s)=1$ and $\sum_{i}\pi_2(a_i|s)=1$, we can easily infer that $\sum_{i}\pi_3(a_i|s)=1$. Therefore, the combined policy $\pi_3$ can be seen as a stochastic policy that integrates two human strategies. The adjustment factor $\alpha$ can be used to modify the tendency behavior of the agent: if $\alpha=0$, $\pi_3$ becomes $\pi_2$ and vice versa. Moreover, the policy $\pi_3$ with $\alpha > 0$ can drastically reduce ``local stuck" phenomenon because of its stochastic behavior.

\begin{table}[!t]
\renewcommand{\arraystretch}{1.3}
\caption{Environment \& Algorithm settings}
\label{table1}
\centering
\begin{tabular}{ccc}
\hline
\hline
Name & Parameter & Value\\
\hline
Breakout   &Episode max steps&10000\\
		   &Number of skipped frames&4\\
		   &Loss of life marks terminal state&No\\
		   &Repeat action probability&0\\
		   &Pixel-wise maximum over&No\\
		   &two consecutive frames&\\
\hline
A3C	       &Optimized learning rate&0.004\\
		   &Discounted factor $\gamma$&0.99\\
		   &Beta&0.1\\	
		   &Number of history frames&4\\
		   &Global normalization clipping&40\\
		   &Asynchronous training steps&5\\
		   &Number of threads&8\\
		   &Anneal learning rate&Yes\\
		   &Optimizer&RMS optimizer\\
		   &RMS's decay&0.99 \\
           &RMS's epsilon&1e-6\\
\hline
\hline
\end{tabular}
\end{table}

This mixed strategy approach can be extendable to $N$ policies ($N > 2$) using the following guidelines:

1) Given a complicated task $T$, we divide $T$ into $N$ smaller tasks $t_i (i=1..N)$ and solve these tasks using $N$ different strategies to achieve the best overall performance.

2) We select a suitable deep RL algorithm (such as A3C) and train $N$ policy networks. Each policy network corresponds with each human strategy.

3) For each policy $\pi_i$, we assign a priority $\alpha_i$ so that $0 \leq \alpha_i \leq 1$ and $\sum_{i}{\alpha_i}=1$. We then infer a combined policy $\pi_N$ using the following generalized version of the $\epsilon$-greedy rule with $N$ strategies, as shown below:

\begin{equation*}
    \pi_N(a|s) = \frac{\epsilon}{|\mathcal{A}|} + \sum_{i}^{N}\alpha_i(1-\epsilon)\pi_i(a|s).
\label{eq:6}
\end{equation*}

\begin{figure}[!t]
\centering
\includegraphics[width=3.4in]{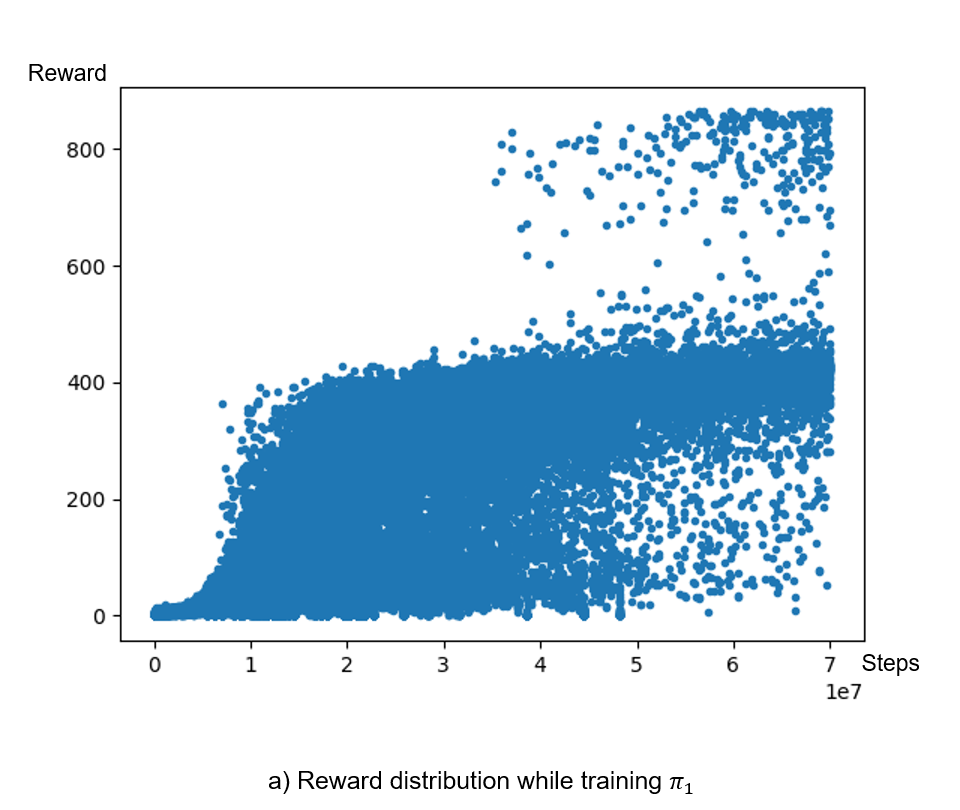}
\includegraphics[width=3.4in]{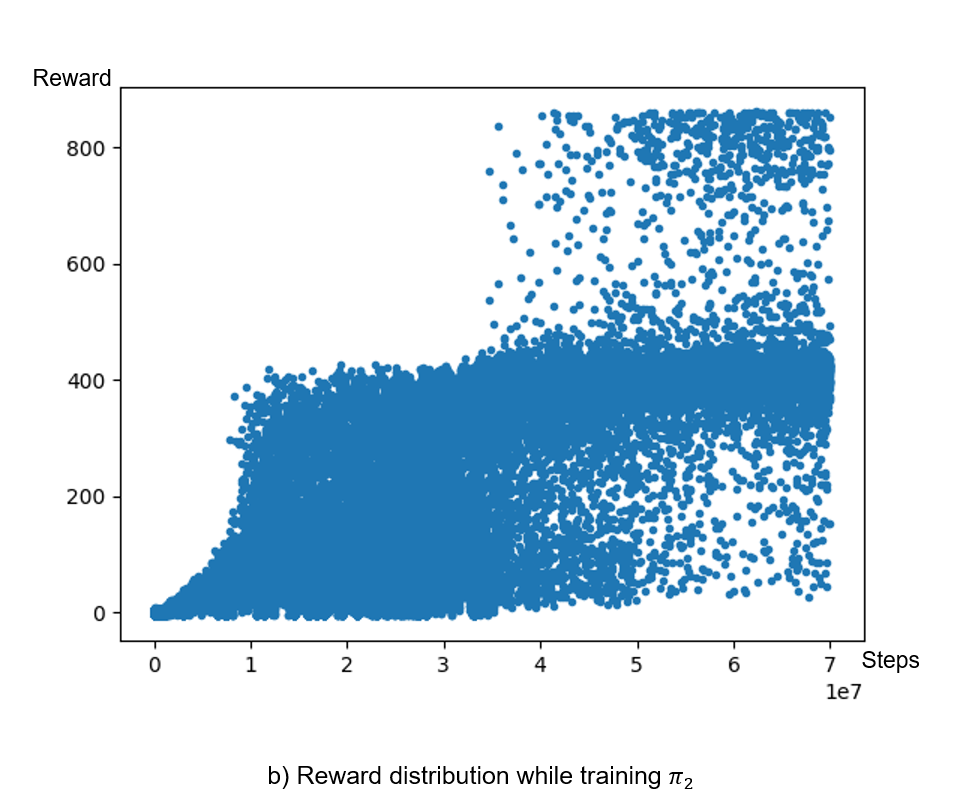}
\caption{Reward distribution while training $\pi_1$ and $\pi_2$.}
\label{fig:5} 
\end{figure}

\begin{figure*}
\begin{multicols}{3}
\includegraphics[width=2.3in]{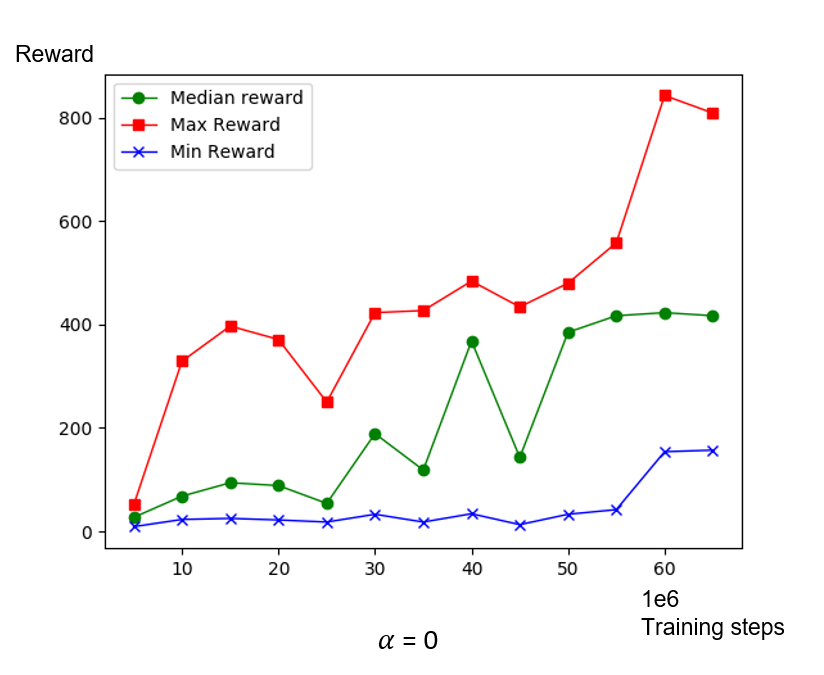}\par
\includegraphics[width=2.3in]{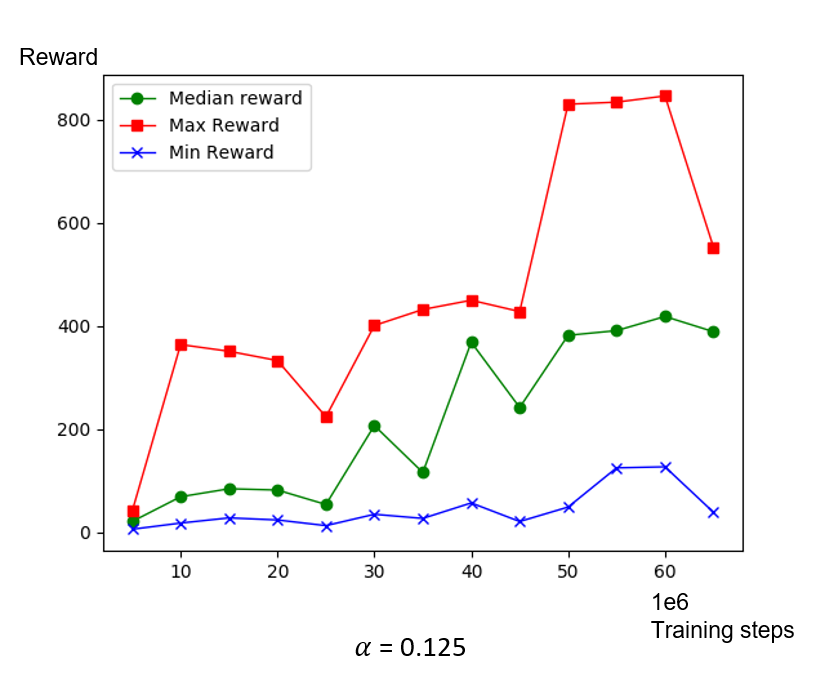}\par
\includegraphics[width=2.3in]{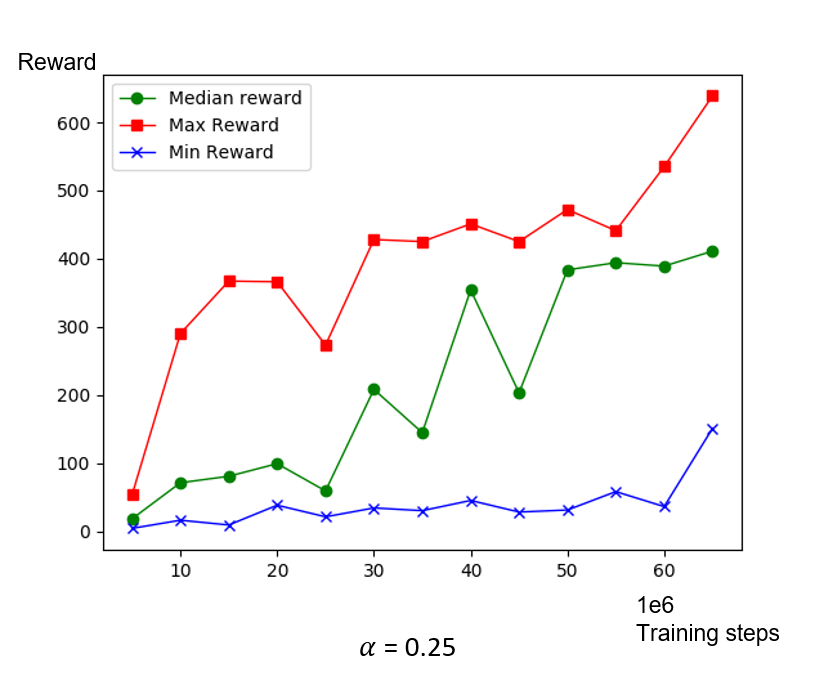}\par
\includegraphics[width=2.3in]{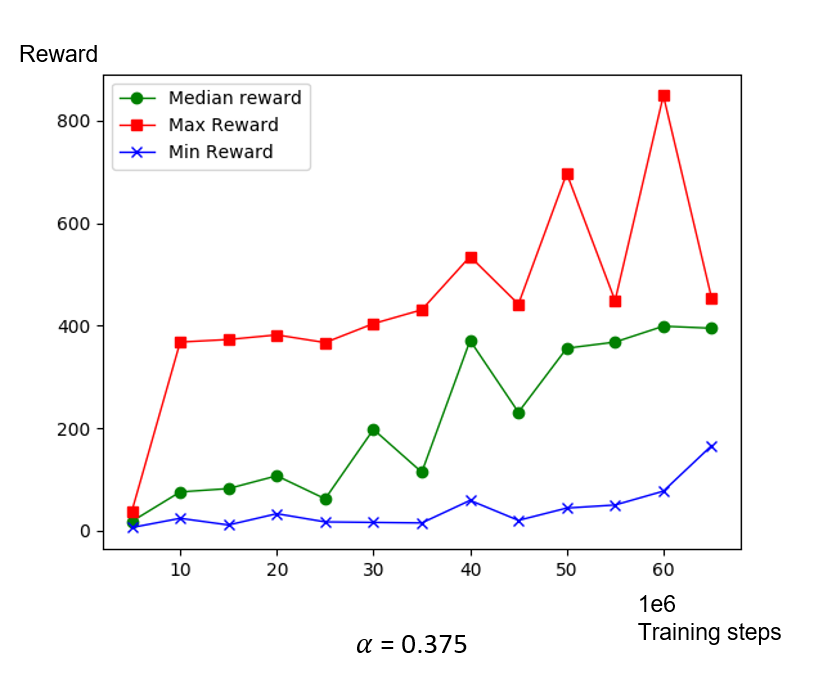}\par
\includegraphics[width=2.3in]{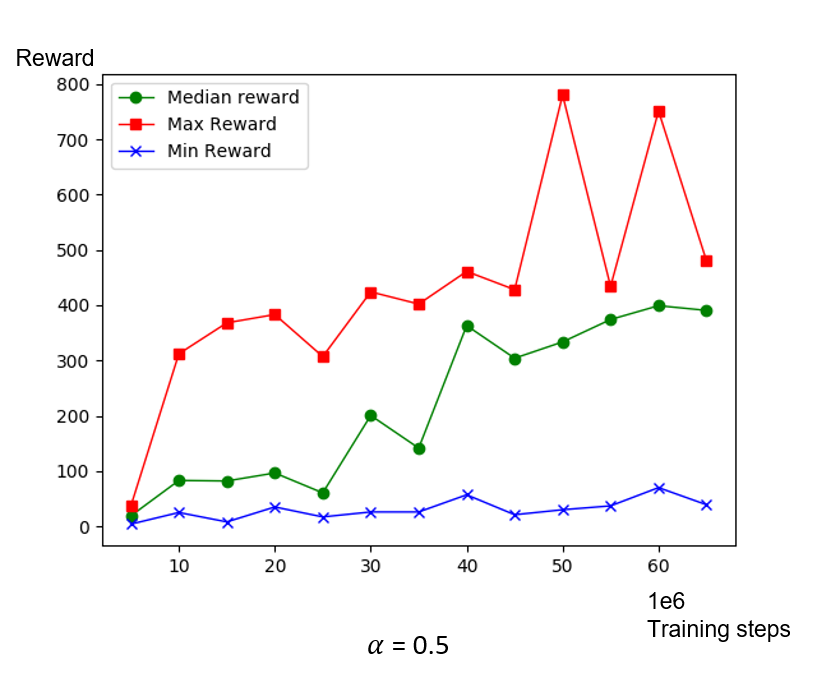}\par
\includegraphics[width=2.3in]{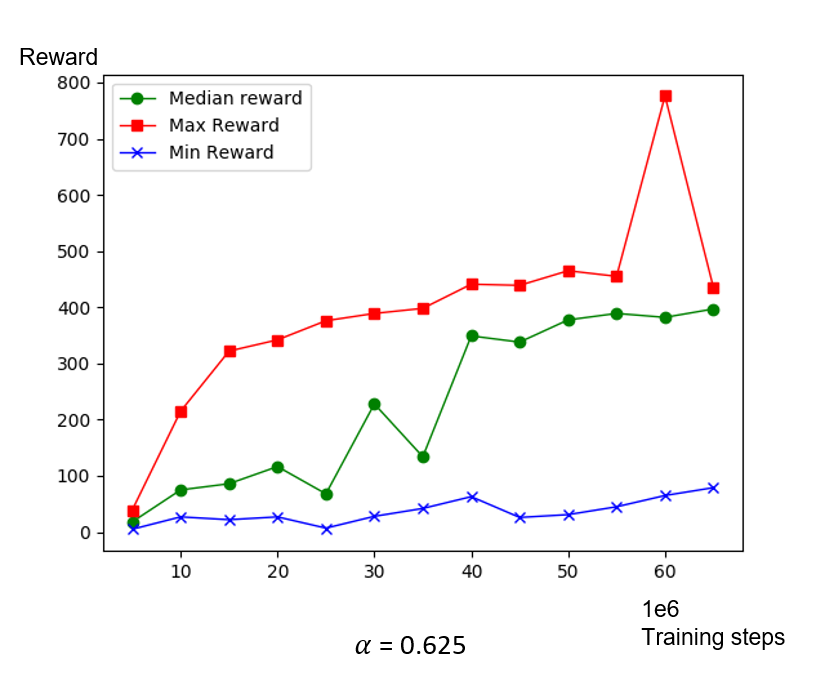}\par
\includegraphics[width=2.3in]{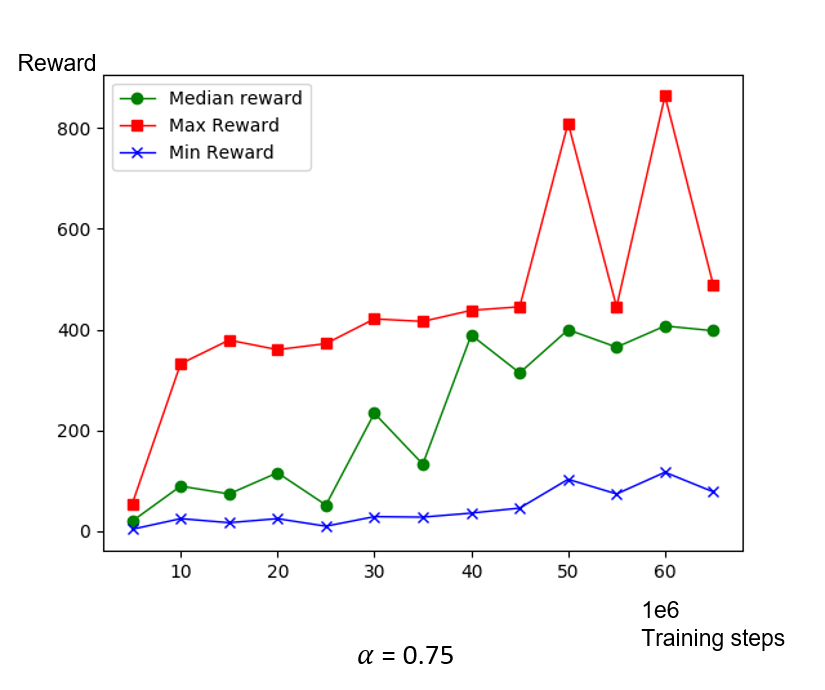}\par
\includegraphics[width=2.3in]{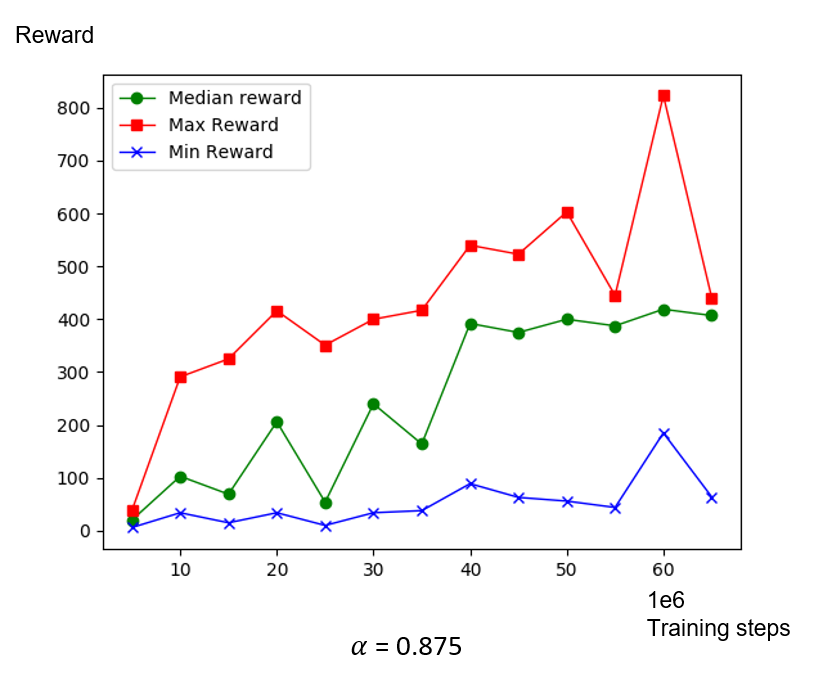}\par
\includegraphics[width=2.3in]{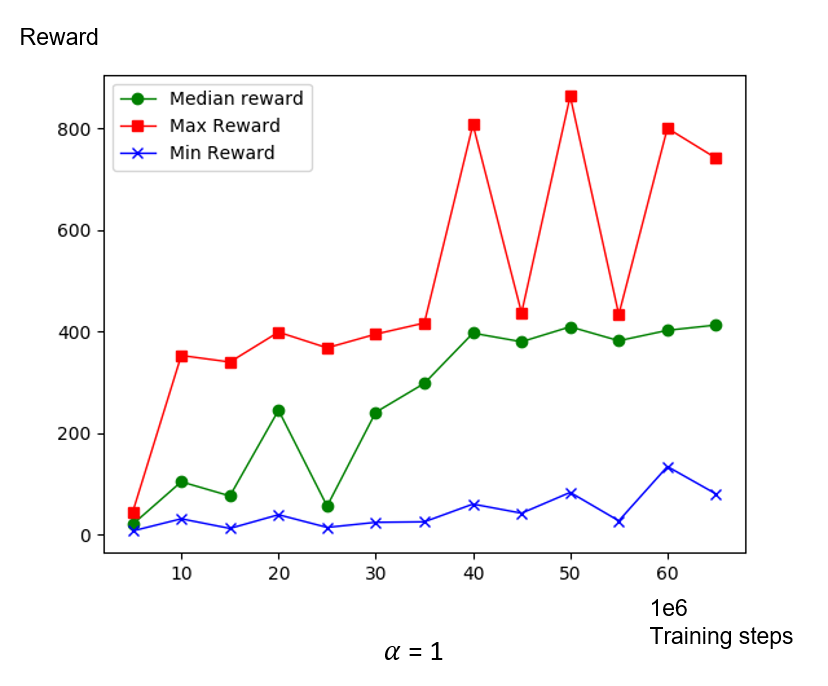}\par
\end{multicols}
\caption{Mixed strategy policy with different value of $\alpha$. We run 120,000 steps and record minimum, maximum, and average median reward for each checkpoint.}
\label{fig:6} 
\end{figure*}

\section{Simulation Results}
\label{sec:4}
In this section, we use the Breakout game as an environment to examine our mixed strategy scheme. We train two different strategies, $\pi_1$ and $\pi_2$, using A3C. Table~\ref{table1} summarizes the setting parameters of Breakout and A3C. The algorithm is run on a computer with a GTX 1080 Ti graphics card. Unlike $\pi_1$, in the training process of $\pi_2$, we allocate a negative reward of $-1$ for each life lost. Moreover, we train both policies in 70 million steps, which is equivalent to 280 million frames of Breakout. 

\begin{figure}[!t]
\centering
\includegraphics[width=3.6in]{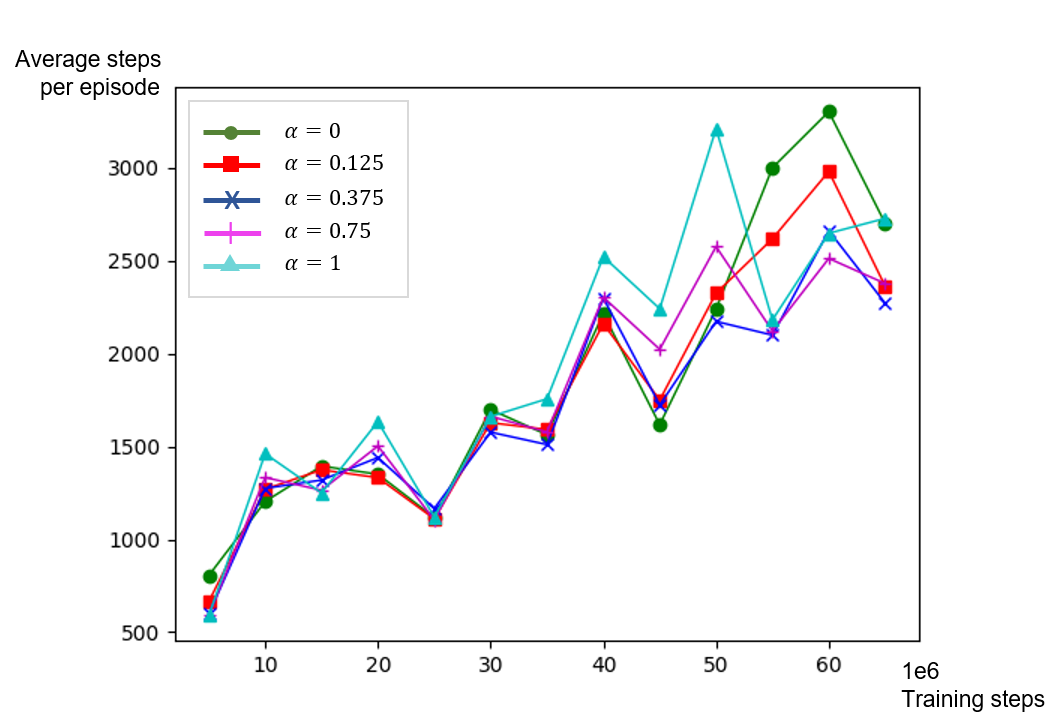}
\caption{Average number of steps in each episode with different values of $\alpha$.}
\label{fig:7} 
\end{figure}

During the training process, we collect the total reward achieved in each episode and record it, as seen in Fig.~\ref{fig:5}. It is evident that in policy $\pi_2$, it is easier to achieve a score of 800 than in $\pi_1$. Therefore, the proposed policy $\pi_2$ achieves a maximum score with a higher probability than that of $\pi_1$. However, the use of pure strategy $\pi_2$ is not recommended because it is prone to the ``local stuck" phenomenon as mentioned in the previous section. Therefore, a mixed strategy is used to balance the maximum score achievement and the average number of steps per episode. Given the same score, it is preferable to use a policy that uses a smaller number of steps. In Fig.~\ref{fig:6}, the adjustment parameter $\alpha$ is used to balance the two strategies. We also keep $\epsilon=0.01$ in all cases. We see that, with $\alpha \leq 0.125$, the probability of achieving a score of 800 at 60 million training steps is high. Therefore, it is desirable to assign $\alpha=0.125$ in the Breakout game. In a real-world application, the choice of $\alpha$ depends on the goal objective, and it can be altered in the real time to adapt with the environment. Finally, Fig.~\ref{fig:7} shows average number of steps used in each episode with different values of $\alpha$. As expected, the pure policy $\pi_2$ ($\alpha=0$) uses the highest number of steps, but can be fixed by increasing $\alpha$. In summary, with $\alpha=0.125$, we obtain a balanced performance between maximum score achievement and average number of steps per episode.

\section{Conclusions}
\label{sec:5}

In this paper, we introduce an extended approach that applies human strategy to deep reinforcement learning. This marks the first step to building a human-like agent that can adapt to its environment using human strategies. Because the limited scope of this paper, we only simulated our mixed strategy approach using the Breakout game, but the proposed scheme can be applied to other Atari games, as well as to real-world problems. The simulation results confirm that our the mixed strategy approach is efficient and promising. We also provide helpful guidelines to solve a complicated task by mimicking the divide-and-conquer strategy of human behaviors. Our future work will continue to work on building human-like agents that can automatically adapt to their environments using different learning strategies. 

\end{document}